\documentclass[10pt,letterpaper]{article}

\usepackage{ccn}
\usepackage{pslatex}
\usepackage{apacite}
\usepackage{graphicx}

\title{Do Vision Transformers See Like Humans? Evaluating their Perceptual Alignment}

\author{{{\large \bf Pablo Hernández-Cámara$^{a*}$}, \bf Jose Manuel Jaén-Lorites$^{b}$, \bf Jorge Vila-Tomás$^{a}$, Valero Laparra$^{a}$, \bf Jesus Malo$^{a}$}\\
$^{a}$ Image Processing Lab, Universidad de Valencia, Paterna, Spain\\
$^{b}$ Center for Biomaterials and Tissue Engineering Universitat Politecnica de Valencia,  Valencia, Spain\\
$^{*}$ Corresponding author: pablo.hernandez-camara@uv.es}
 
\begin{document}

\maketitle

\section{Abstract}
{
\bf
Vision Transformers (ViTs) achieve remarkable performance in image recognition tasks, yet their alignment with human perception remains largely unexplored. This study systematically analyzes how model size, dataset size, data augmentation and regularization impact ViT perceptual alignment with human judgments on the TID2013 dataset.
Our findings confirm that larger models exhibit lower perceptual alignment, consistent with previous works. Increasing dataset diversity has a minimal impact, but exposing models to the same images more times reduces alignment. Stronger data augmentation and regularization further decrease alignment, especially in models exposed to repeated training cycles. These results highlight a trade-off between model complexity, training strategies, and alignment with human perception, raising important considerations for applications requiring human-like visual understanding.
}
\begin{quote}
\small
\textbf{Keywords:} 
Vision Transformer; Human Alignment; Perceptual Similarity; Human Visual Perception; Image Quality
\end{quote}

\section{Introduction}

Vision Transformers (ViTs) have demonstrated exceptional performance across diverse image recognition tasks \cite{vit}. However, while these models achieve high accuracy on benchmark datasets, their alignment with human perception remains an open question. Understanding how architecture and training strategies influence perceptual alignment is crucial for building more interpretable, robust, capable-to-generalize and human-aligned models. This could benefit applications such as image quality assessment, content generation, and explainable AI.

Prior research has shown that convolutional neural networks (CNNs) can exhibit varying degrees of perceptual alignment based on their architecture and training settings \cite{zhang2018unreasonable, kumar2022better, hernandez2025dissecting}. However, ViTs rely on self-attention mechanisms, enabling global context modeling that may alter their perceptual representations \cite{raghu2021vision}. While some studies have examined perceptual alignment in CNNs, a systematic analysis of ViTs under different training conditions is still limited \cite{kumar2022better, hernandez2024measuring}.

In this study, we investigate how key factors such as model size, dataset size, data augmentation, and regularization influence the perceptual alignment of ViTs. Using a collection of pretrained ViTs, we evaluate their outputs on the TID2013 dataset \cite{ponomarenko2015image}, a widely used benchmark that captures human visual quality assessments. It contains a diverse set of images with controlled distortions, making it an effective tool for measuring perceptual similarity between human observers and computational models.

\section{Methods}

\subsection{Models and Factors}

We analyze a collection of ViTs trained for image classification \cite{steiner2021train}. This collection includes over 50,000 pretrained models trained under different configurations, allowing us to systematically explore how variations in architecture and training parameters affect perceptual alignment. Particularly, we focus on the factors varied in this collection of models, such as the model size (small vs. large architecture), the dataset size (number of unique images used during training), samples seen (how many times each image is seen during the training), intensity of data augmentation (strength of augmentations applied) and regularization (effect of techniques such as dropout and stochastic depth).

\subsection{Perceptual Alignment Evaluation}

We evaluate perceptual alignment using the TID2013 dataset \cite{ponomarenko2015image}. It contains 25 original images and 24 distortions at five intensity levels, each pair rated by human observers with the mean opinion score (MOS) reflecting perceptual quality, i.e. how much humans see the difference for each particular image pair. For each image pair, we first pass both the original and distorted images through the ViT and extract the encoder outputs for each image. Then, we compute the Euclidean distance between the encoder outputs of the original and distorted images, which represents the model's perceptual dissimilarity. We compute the Spearman correlation between the model distances and the human MOS across all image pairs, a quantitative measure of the model's alignment with human perceptual judgments. By applying this procedure across multiple ViTs with varying training factors, we systematically assess how these factors influence the perceptual alignment of Vision Transformers.

\section{Results}

\begin{figure}[ht]
\begin{center}
\includegraphics[width=0.48\textwidth]{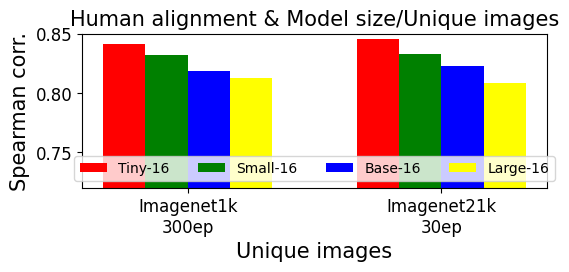}
\end{center}
\vspace{-0.6cm}
\caption{ViT perceptual alignment on TID2013 depending on model size (colors) and the dataset size for a fixed compute, i.e. same quantity of images seen.} 
\label{fig_results_modelsize_datasetsize}
\end{figure}

\textbf{Model Size and Dataset Size:} As shown in Figure \ref{fig_results_modelsize_datasetsize}, larger models exhibit lower perceptual alignment across all dataset configurations. Additionally, increasing dataset diversity ( from 1.3M to 13M unique images) has minimal impact when the total number of images seen remains constant. This suggests that model size influences perceptual properties more than dataset diversity.

\begin{figure}[ht]
\begin{center}
\includegraphics[width=0.48\textwidth]{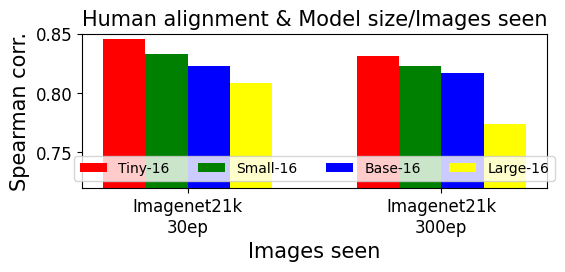}
\end{center}
\vspace{-0.6cm}
\caption{ViT perceptual alignment with TID2013 depending on model size (colors) and the number of images seen for a fixed dataset size, i.e. same number of unique images.\vspace{-0.2cm}} 
\label{fig_results_modelsize_imagesseen}
\end{figure}

\textbf{Exposure Frequency:} Figure \ref{fig_results_modelsize_imagesseen} illustrates the effect of image exposure frequency, i.e. the number of times each image is seen during the training. Models that see each image more times during training show lower perceptual alignment, i.e. the models suffer from overfitting. This effect is most pronounced in the larger models (24-layer ViT), aligning with prior findings that prolonged training and, therefore, more accuracy on the training objective can reduce alignment with human perception \cite{gomez2020color, li2022contrast,  kumar2022better, hernandez2025dissecting}.

\begin{figure}[ht]
\begin{center}
\includegraphics[width=0.48\textwidth]{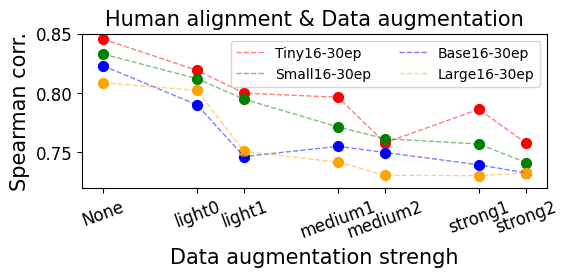}
\end{center}
\vspace{-0.6cm}
\caption{ViT perceptual alignment with TID2013 depending on the data augmentation (x-axis).
\vspace{-0.2cm}} 
\label{fig_dataaugmentation}
\end{figure}

\begin{figure}[ht]
\begin{center}
\includegraphics[width=0.48\textwidth]{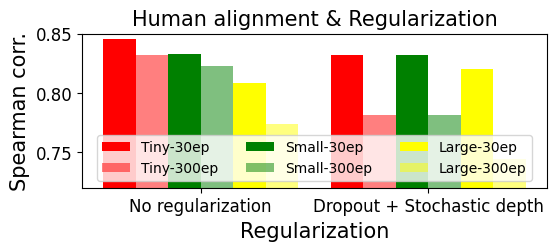}
\end{center}
\vspace{-0.6cm}
\caption{ViT perceptual alignment with TID2013 depending on the regularization (x-axis).
\vspace{-0.5cm}} 
\label{fig_regularization}
\end{figure}

\textbf{Data Augmentation and Regularization:} Figure \ref{fig_dataaugmentation} shows that stronger data augmentation consistently decreases perceptual alignment for all model sizes. Similarly, Figure \ref{fig_regularization} demonstrates that regularization also reduces alignment, particularly for the models with extensive training, i.e. the ones that have seen each image more times. This suggests that techniques commonly used to enhance model generalization may inadvertently push ViTs further from human-like perceptual representations.

Finally, figure \ref{fig_regularization} shows how regularization changes the ViT perceptual alignment. It shows that regularization also results in less perceptually aligned models. The models are less aligned for the models trained during more epochs and therefore more overfitted.

\begin{figure}[b]
\begin{center}
\includegraphics[width=0.48\textwidth]{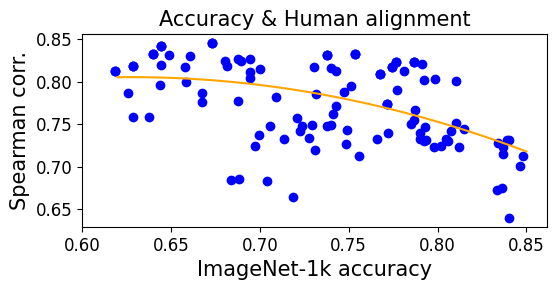}
\end{center}
\vspace{-0.6cm}
\caption{Relation between perceptual alignment and ImageNet-1k classification accuracy with the best fit.\vspace{-0.5cm}} 
\label{fig_inverted_U}
\end{figure}

\section{Conclusions}

This study reveals a trade-off between optimizing ViTs for classification performance and maintaining perceptual alignment with humans. Factors that typically improve classification accuracy—larger models, extended training, stronger data augmentation, and regularization—consistently reduce alignment with human perception. Consistent with previous works, we show in Figure \ref{fig_inverted_U} that more capable models in terms of their accuracy in classification are consistently less human-like \cite{gomez2020color, li2022contrast,  kumar2022better, hernandez2025dissecting}.
These findings suggest that optimizing ViTs purely for task performance may lead to representations that diverge from human visual processing.

Future research should explore ways to balance perceptual alignment and task optimization, particularly for applications requiring human-like interpretation, such as medical imaging, artistic rendering, and explainable AI systems. Understanding how to train AI models that "see" more like humans could lead to more intuitive and reliable machine vision applications.

\section{Acknowledgments}

This work was supported in part by MCIN/AEI/FEDER/UE under Grants PID2020-118071GB-I00 and PID2023-152133NB-I00, by Spanish MIU under Grant FPU21/02256 and in part by Generalitat Valenciana under Projects GV/2021/074, CIPROM/2021/056, and by the grant BBVA Foundations of Science program: Maths, Stats, Comp. Sci. and AI (VIS4NN). Some computer resources were provided by Artemisa, funded by the EU ERDF through the Instituto de Física Corpuscular, IFIC (CSIC-UV).

\bibliographystyle{ccn_style}

\setlength{\bibleftmargin}{.125in}
\setlength{\bibindent}{-\bibleftmargin}

\bibliography{ccn_style}

\end{document}